\documentclass[11pt]{article}
\usepackage{fullpage}
\usepackage{fullpage,xspace,amsmath,amsfonts,amssymb}
\usepackage{times,mathptmx}
\usepackage{graphicx,color}
\usepackage{amsfonts,amsmath,color,float,graphicx,verbatim, times}
\usepackage{amsthm}
\usepackage{algorithm}
\usepackage{algpseudocode}

\newcommand{\sgn}{\mathop{\mathrm{sgn}}}

\theoremstyle{plain}

\newtheorem{definition}{Definition}

\newtheorem{lemma}{Lemma}

\numberwithin{equation}{section}
\newif\ifiscolor
\iscolorfalse
\begin{document}
\pagestyle{empty}
\title{Accurate Streaming Support Vector Machines}
\author{Vikram Nathan \and Sharath Raghvendra}
\date{}
\maketitle
\thispagestyle{empty}

\section{Abstract}

A widely-used tool for binary classification is the Support Vector Machine (SVM), a supervised learning technique that finds the ``maximum margin'' linear separator between the two classes. While SVMs have been well studied in the batch (offline) setting, there is considerably less work on the streaming (online) setting, which requires only a single pass over the data using sub-linear space. Existing streaming algorithms are not yet competitive with the batch implementation. In this paper, we use the formulation of the SVM as a minimum enclosing ball (MEB) problem to provide a streaming SVM algorithm based off of the blurred ball cover originally proposed by Agarwal and Sharathkumar. Our implementation consistently outperforms existing streaming SVM approaches and provides higher accuracies than libSVM on several datasets, thus making it competitive with the standard SVM batch implementation.

\section{Introduction}

Learning and classification with a large amount of data raises the need for algorithms that scale well in time and space usage with the number of data points being trained on. \textit{Streaming} algorithms have properties that do just that: they run in a single pass over the data and use space polylogarithmic in the total number of points. The technique of making a single pass over the data has three key advantages: 1) points may be seen once and then discarded so they do not take up additional storage space; 2) the running time scales linearly in the size of the input, a practical necessity for data sets with sizes in the millions, and 3) it enables these algorithms to function in a streaming model, where instead of data is not immediately available, individual data points may arrive slowly over time. This third feature enables data to be learned and models to be updated "online" in real time, instead of periodically running a batch update over all existing data.

Support Vector Machines (SVMs) are one such learning model that would benefit from an efficient and accurate streaming representation. Standard 2-class SVMs models attempt to find the maximum-margin linear separator (i.e. hyperplane) between positive and negative instances and as such, they have a very small hypothesis complexity but provable generalization bounds \cite{vapnik}. There have been several implementations of a streaming SVM classifier (\cite{lasvm}, \cite{streamsvm}, \cite{perceptron}), but so the most effective version has been based off the reduction from SVM to the Minimum Enclosing Ball (MEB) problem introduced by~\cite{cvm}. The connection between these two problems has made it possible to harness the work done on streaming MEBs and apply them to SVMs, as was done in~\cite{streamsvm}. In this paper, we utilize the Blurred Ball cover approximation to the MEB problem proposed in~\cite{bbcover} to obtain a streaming SVM that is both fast and space efficient. We also show that our implementation not only outperforms existing streaming SVM implementations (including those not based off of MEB reductions) but also that our error rates are competitive with LibSVM, a state-of-the-art batch SVM open-source project available [here]. 

\section{Background}
\label{background}

The Core Vector Machine (CVM) was introduced by~\cite{cvm} as an new take on the standard Support Vector Machine (SVM). Instead of attempting to solve a quadratic system, the CVM makes use of the observation that many common kernels for the standard SVM can be viewed as Minimum Enclosing Ball (MEB) problems. Consider the following SVM optimization problem on $m$ inputs $(\mathbf{x}_i, y_i)$:
\begin{align}
\label{reduction}
\min_{\mathbf{w}, b, \rho, \xi_i} &\|\mathbf{w}\|^2 + b^2 - 2\rho + C\sum_{i = 1}^n \xi_i^2 \\
s.t. \quad& y_i(\mathbf{w} \cdot \mathbf{\varphi}(\mathbf{x}_i) + b) \geq \rho - \xi_i\quad i = 1, \ldots, n \nonumber
\end{align}
where $\mathbf{x}_i$ and $y_i$ are the data points and labels, respectively, and $\phi$ is a feature map induced by the kernel of choice. Here, the $\xi_i$ are error cushions that specify the cost of misclassifying $\mathbf{x}_i$. Let $\mathbf{w}^*$ be the optimal separating hyperplane. \cite{cvm} showed that if the kernel $K(\mathbf{x}, \mathbf{y}) = \mathbf{\varphi}(\mathbf{x}) \cdot \mathbf{\varphi}(\mathbf{y})$ satisfies $K(\mathbf{x}, \mathbf{x}) = \kappa$, a constant, then $\mathbf{w}^*$ can be found by finding the minimum enclosing ball of the points $\{\mathbf{\overline{\varphi}}(\mathbf{x}_i, y_i)\}$, where
\[ \overline{\varphi}_i(\mathbf{x}_i, y_i) = \left[\begin{array}{c} y_i\mathbf{\varphi}(\mathbf{x}_i) \\ y_i \\ \frac{1}{\sqrt{C}} \mathbf{e}_i \end{array}\right] \]
where $\mathbf{e}_i$ is the $i$th standard basis element (all zeroes except for the $i$th position, which is 1). If $\mathbf{c}^*$ is the optimal MEB center, then $\mathbf{w}^* = \mathbf{c}^* \cdot \left(\mathbf{e}_1 + \ldots + \mathbf{e}_m\right)$.

A couple of things to note about this equivalence: first, it transforms the supervised SVM training problem into an unsupervised one - the MEB is blind to the true label of the point. Second, the notion of a \emph{margin} in the original SVM problem is transformed into a \emph{core set}, a subset of inputs such that finding the MEB of the corset is a good approximation to finding the MEB over the entire input. As such, core sets can be thought of as the minimal amount of information that defines the MEB. The implementation of the CVM in~\cite{cvm} follows the MEB approximation algorithm described in~\cite{badoiu}: given $\varepsilon > 0$, add to the core set $C$ the point $\mathbf{z}$ that is the farthest from the current MEB center $c$. Recompute the MEB from the core set and continue until all points are within $(1+\varepsilon)R$ from $c$, where $R$ is the radius of the MEB.

The core vector machine achieves a $1+\epsilon$ approximation factor but makes $|C|$ passes over the data and requires storage space linear in the size of the input, an approach that doesn't scale for streaming applications. To this end, \cite{streamSVM} presented the StreamSVM, a streaming version of the CVM, which computes the MEB over the input $\{\overline{\mathbf{\varphi}}_i\}$ in a streaming fashion, keeping a running approximate MEB of all the data seen so far and adjusting it upon receiving a new input point. The StreamSVM used only constant space and using a small lookahead resulted in a favorable performance compared to libSVM (batch) as well as the streaming Perceptron, Pegasos, and LASVM implementations. However, the streaming MEB algorithm that powers StreamSVM is only approximate and offers a worst-case approximation ratio of between $\frac{1 + \sqrt{2}}{2}$ and $\frac{3}{2}$ of the true MEB, leaving open the possibility of a better streaming algorithm to improve the performance of StreamSVM.

In this paper, we present the Blurred Ball SVM, a streaming algorithm based on the blurred ball cover proposed in~\cite{bbcover}. It takes a parameter $\varepsilon > 0$ and keeps track of multiple core sets (and their corresponding MEBs), performing updates only when an incoming data point lies outside the union of the $1+\varepsilon$ expansion of all the maintained MEBs. The Blurred Ball SVM also makes a single pass over the input, with space complexity independent of $n$.

\section{Algorithm}

The algorithm consists of two parts: a \emph{training} procedure to update the blurred ball cover and a \emph{classification} method, which harnesses the blurred ball to label the point with one of the two possible classes. For simplicity, we choose the classes to be $\pm 1$. 

As described above, a ball with radius $r$ and center $c$ is a linear classifier consisting of a hyperplane passing through the origin with normal $c$. For the rest of this paper, we will require the following assumptions, established by Tsang et al:
\begin{itemize}
\item The data points $\overline{\phi}(\mathbf{x}_i, y_i)$, denoted by $D$, are linearly separable (this is always the case if $C < \infty$).
\item $|\overline{\phi}(\mathbf{x}_i, y_i)| = \kappa$, a constant.
\end{itemize}
With these assumptions, the training procedure is described in Algorithm~\ref{updatealgo} and is identical to the blurred ball cover update described in Algorithm 1 of~\cite{blurredball}. 

\begin{algorithm}
\begin{algorithmic}[1]
\Function {Train}{$x_i, y_i, L$}
\State $\textit{cores} \gets \text{[]}$
\State Compute $x' = \overline{\varphi}(x_i, y_i)$ (normalized to norm $\kappa$) as defined above
\State Add $x'$ to the lookahead buffer \textit{buf}
\If {\textit{buf} $< L$} \Return
\EndIf
\If {$\exists x' \in B$ s.t. $x'$ is not in the $1+\varepsilon$ expansion of any MEB in \textit{cores}}
\State $c, B \gets \text{new core set, MEB of $\{B\} \; \cup $ \textit{cores}}$
\State Discard any core set with MEB radius smaller than $r(B) \cdot \varepsilon/4$
\State $\textit{cores} \gets \textit{cores } \cup (c,B)$
\EndIf
\State \textit{buf} $\gets \{\}$
\EndFunction
\end{algorithmic}
\caption{Outline of training procedure for the Blurred Ball SVM}
\label{updatealgo}
\end{algorithm}

For the purposes of analysis, we show the following properties of the linear classifier that results from the blurred balls:
\begin{lemma}
\label{mebequivsvm}
A ball $B$ with center $c$ and radius $r$ corresponds to a linear classifier with hyperplane $h$ having the following properties:
\begin{enumerate}
\item $|c| > 0$ and $r < \kappa$.
\item Its margin has size $\sqrt{\kappa^2 - r^2}$.
\item A point $p$ lies inside $B$ iff $(p - c) \cdot c \geq 0$, with equality for support vectors, which lie on $\partial B$.
\end{enumerate}
\begin{proof}
First, note that $|c| > 0$. Suppose instead that $|c| = 0$. Then we use the following property of a MEB: any half-space $H$ such that $c \in \partial H$ contains at least one data point used to construct the MEB. Suppose that $r = \kappa$, i.e. $|c| = 0$. Then this property shows that there is no hyperplane passing through the origin that contains all points entirely on one side.
Now, assume that the data points are separable and let $\mathbf{h}$ be the normal of the hyperplane that separates the raw data points $\mathbf{x}_i$ such that $\mathbf{x}_i \cdot \mathbf{h} > 0$ for positively labeled points and $< 0$ for negatively classified points. Then, $\overline{\mathbf{\phi}}_i \cdot \mathbf{h} = y_i\mathbf{x}_i \cdot \mathbf{h} > 0$ for all $i$, a contradiction to the above property if $|c| = 0$. We can thus assume that $|c| > 0$ and $r < \kappa$ for a linearly separable dataset.

The reduction described in Section~\ref{background} shows that the linear separator defined by a MEB with center $\mathbf{c}$ and radius $r$ is a hyperplane with normal parallel to $\mathbf{c}$. Let $\mathbf{d}$ be the point farthest from the origin such that $(\mathbf{p} - \mathbf{d}) \cdot \mathbf{d} \geq 0$ for all data points $\mathbf{p}$. In other words, the maximum margin is $\mathbf{d}$ and $\mathbf{c} \| \mathbf{d}$ from the reduction. We can further conclude that $\mathbf{c} = \mathbf{d}$ as follows: let $S = \{\mathbf{p} | (\mathbf{p} - \mathbf{c}) \cdot \mathbf{c} = 0\}$ denote the support vectors, those that lie on the margin and are a distance $\|\mathbf{d}\|$ from the maximum-margin hyperplane. The ball $B(\mathbf{d}, \|\mathbf{s} - \mathbf{d}\|)$ for $\mathbf{s} \in S$ includes all data points (it intersects $B(\mathbf{0}, \kappa)$ along the margin).
If $\mathbf{c} \neq \mathbf{d}$, then $\text{argmax}_{p \in D} \|p - c\|^2 = \|p - d\|^2 + \|d - c\|^2 > \|s - d\|$ and $c$ is thus not the center of the MEB (since $d$ is strictly better). So $\mathbf{c} = \mathbf{d}$ and the MEB has radius $\|s - c\| = |c|^2 + \kappa^2$ for $s \in S$. Therefore, the margin is $\|c\| = \sqrt{\kappa^2 - r^2}$.

Since $B^* = B(\mathbf{c}, \|\mathbf{s} - \mathbf{c}\|)$ intersects $B(\mathbf{0}, \kappa)$ only along the hyperplane that defines the margin, $S \subset \partial B^*$, and $D \setminus S \subset B^* \setminus \partial B^*$.

\end{proof}
\end{lemma}

Since we have multiple linear separators, we have the ability to combine them in a non-linear fashion to classify a new point.
\begin{definition} Define the \emph{support} of a point $p$ to be $\text{Sup}(p) = \{B \in \text{cores} | p \in B\}$, the cores in the blurred ball cover that contain $p$.
\end{definition}

\begin{definition} Define the \emph{score} of a point $p$ to be:
\[ S(p) = \sum_{B \in \text{Sup}(p)} p \cdot \frac{c_B}{\|c_B\|}, \]
the sum of the distances of $p$ to the separator of all the classifiers containing $p$. 
\end{definition}
Note that $\text{Sup}(p) \cap \text{Sup}(-p) = \emptyset$, since each ball has $r < \kappa$.

\begin{algorithm}
\begin{algorithmic}[1]
\Function {Classify with Majority}{$x_i$}
\State \Return $H(p) = \sgn\left[S(p) - S(-p)\right]$
\EndFunction
\end{algorithmic}
\caption{Example classification procedure.}
\label{classifyalgo}
\end{algorithm}

\section{Results}

\begin{figure}[htbp]
\centering
\includegraphics[scale=0.6]{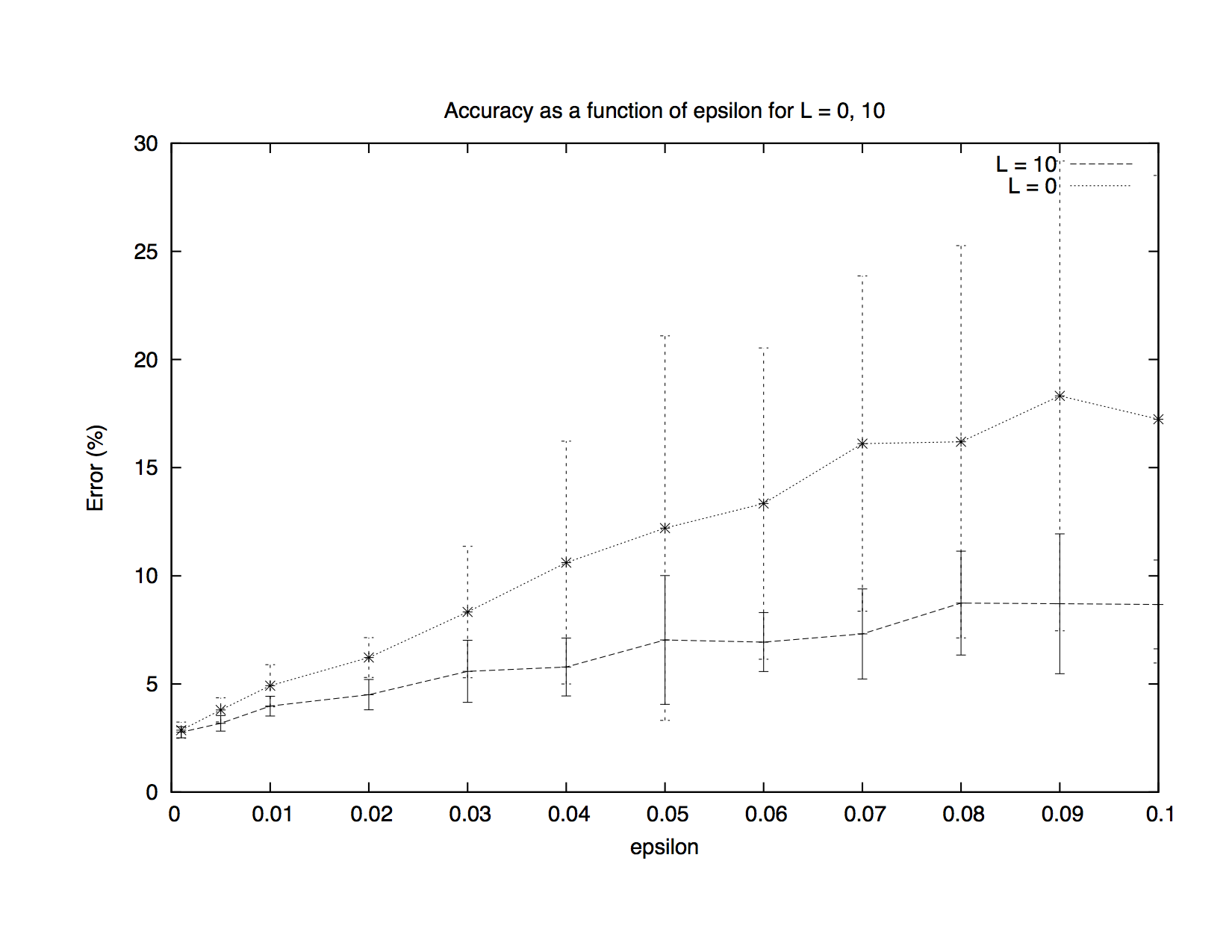}
\caption{Error as a function of $\varepsilon$, with lookaheads $L = 0$ and $L = 10$. Despite diverging for large $\varepsilon$, the accuracies with both lookaheads were much more similar for small $\varepsilon$.}
\label{lookaheadgraph}
\end{figure}

\begin{figure}[htbp]
\centering
\includegraphics[scale=0.6]{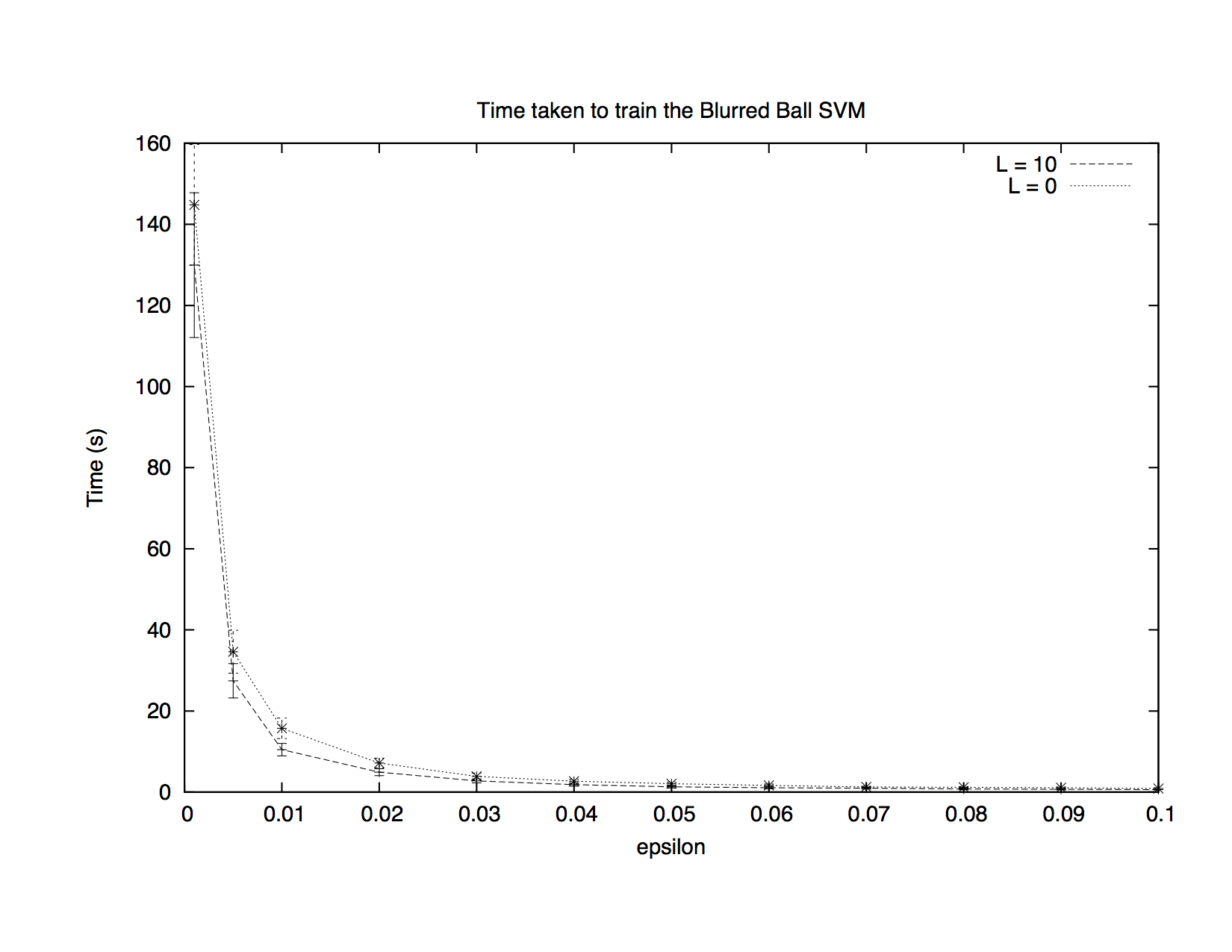}
\caption{Time taken as a function of $\varepsilon$, with lookaheads $L = 0$ and $L = 10$.}
\label{lookaheadgraph-time}
\end{figure}

We ran the Blurred Ball SVM on several canonical datasets and compared the accuracy of each run with the batch LibSVM implementation, the Stream SVM proposed by Subramanian, and the streaming setting of the Perceptron (which runs through the data only once but is otherwise identical to the perceptron training algorithm). Table~\ref{results} shows the experimental results. All trials were averaged over 20 runs with respect to random orderings of the input data stream. The Perceptron, LASVM and Stream SVM data were taken from the experiments documented in~\cite{streamsvm}. The Blurred Ball SVM on the MNIST dataset was run with $\varepsilon = 0.001$ and $C = \infty$, and on the IJCNN dataset was run with $\varepsilon = 10^{-6}$ and $C = 10^5$. The choice of $\varepsilon$ and $C$ was determined coarsely through experimentation. We offer two versions of the Blurred Ball SVM - using lookahead buffer sizes of $L = 0$ and $L = 10$. Figures~\ref{lookaheadgraph} and~\ref{lookaheadgraph-time} compare performance of different lookaheads as $\varepsilon$ is varied. All experiments were run on a Macintosh laptop with a 1.7 GHz processor with 4 GB 1600 MHz standard flash memory. 

It's clear that the Blurred Ball SVM outperforms other streaming SVMs, but even more surprising is that it also manages to outperform the batch implementation on the MNIST dataset. We suspect that this is due to the fact that our classifier allows for non-convex separators.

\begin{table}[htbp]
\centering
\begin{tabular}{c|c|c|c|c}
 & MNIST (0 vs 1) & MNIST (8 vs 9)  & IJCNN & w3a \\ \hline
Dim & 784 & 784 & 22 & 300  \\ 
Train & 12665 & 11800 & 35000 & 44837 \\
Test & 2115 & 1983 & 91701 & 4912 \\ \hline
\bf{LibSVM} & 99.52 & 96.57 & 91.64 & 98.29 \\ \hline
\bf{Perceptron} & 99.47 & 95.9 & 64.82 & 89.27 \\ \hline
\bf{LASVM} & 98.82 & 90.32 & 74.27 & 96.95 \\ \hline
\textbf{StreamSVM} (L = 10) & 99.71 & 94.7 & 87.81 & 89.06 \\ \hline
\textbf{Blurred Ball SVM} (L = 0) & 99.89 & 97.14 & 89.64 & \textbf{97.14}  \\
\textbf{Blurred Ball SVM} (L = 10) & \textbf{99.93}  & \textbf{97.23} & \textbf{90.82} & 97.08 \\
\end{tabular}
\label{results}
\caption{Results on standard datasets comparing the performance of the Blurred Ball SVM with other streaming SVMs and the batch libSVM baseline. $L$ is the size of the lookahead used in the streaming algorithms. Measurements were averaged over 20 runs (w.r.t random orderings of the input stream). The bold number for each dataset is the streaming SVM that gave the highest accuracy for that dataset.}
\end{table}

\section{Further Work}

Being able to learn an SVM model in an online setting opens up myriad possibilities in the analysis of large amounts of data. There are several open questions whose answers may shed light on a streaming approach with higher accuracy than the Blurred Ball SVM presented here:
\begin{enumerate}
\item Is there a streaming algorithm for maintaining an MEB with better guarantees than the Blurred Ball cover proposed by~\cite{bbcover}? The paper originally provided a bound of $\frac{1+\sqrt{3}}{2} \approx 1.3661$, which was improved by~\cite{chan} to less than $1.22$. Although \cite{bbcover} showed that it is impossible to achieve an arbitrarily small approximation factor, with $1 + \epsilon$ for any $\epsilon > 0$, it's possible that a better streaming MEB algorithm exists with provable bounds better than the 1.22 factor demonstrated by~\cite{chan}.
\item The structure of the points in this SVM setup is unique: all data points lie on a sphere of radius $\kappa$ centered at the origin. Although there is no streaming MEB algorithm for unrestricted points, does this specific structure lend itself to a $1+\epsilon$ MEB approximation? If so, we would be able to construct an SVM with separator arbitrarily close to the optimal.
\end{enumerate}

\section{Conclusion}

We have presented a streaming, or ``online'' algorithm for SVM learning by making use of a reduction from the Minimum Enclosing Ball problem. Our training algorithm is tunable using the $\varepsilon$ parameter to adjust the desired approximation ratio. We also came up with multiple types of classifiers, some of them non-convex, and showed that our implementation surpassed the accuracy of other streaming implementations. One surprising finding is that our implementation surpasses the standard libSVM dataset on canonical MNIST binary digit classification datasets. Tests on other digit recognition datasets show similar results, suggesting that this better performance could be due to structural idiosyncrasies of the data.

\end{document}